\documentclass[letterpaper, 10 pt, conference]{ieeeconf}
\usepackage[latin1]{inputenc}
\usepackage{amsfonts}
\usepackage{amsmath}
\usepackage[T1]{fontenc}
\usepackage[dvips]{graphicx}
\usepackage{psfrag}
\usepackage{cite}
\usepackage{color}
\IEEEoverridecommandlockouts
\title{\LARGE \bf
Singular surfaces and cusps in symmetric planar 3-RPR manipulators
}

\author{Michel Coste, Philippe Wenger and Damien Chablat
\thanks{The research work reported here was made possible by SiRoPa ANR Project.}
\thanks{M. Coste is with Institut de Recherche Mathématique de Rennes, Université de Rennes I, Campus de Beaulieu, 35042 Rennes, France
        {\tt\small michel.coste@univ-rennes1.fr}}%
\thanks{P. Wenger and D. Chablat are with Institut de Recherche en Communications et Cybern\'etique de Nantes, 1 rue de la no\"e, 44321 Nantes, France
        {\tt\small Philippe.Wenger@irccyn.ec-nantes.fr}, {\tt\small Damien.Chablat@irccyn.ec-nantes.fr}}%
}

\begin{document}

\maketitle
\thispagestyle{empty}
\pagestyle{empty}

\begin{abstract}
We study in this paper a class of 3-RPR manipulators for which the direct kinematic problem (DKP) is split into a cubic problem followed by a quadratic one. These manipulators are geometrically characterized by the fact that the moving triangle is the image of the base triangle by an indirect isometry. We introduce a specific coordinate system adapted to this geometric feature and which is also well adapted to the splitting of the DKP. This allows us to obtain easily precise
descriptions of the singularities and of the cusp edges. These latter second order singularities are important for nonsingular assembly mode changing. We show how to sort assembly modes and use this sorting for motion planning in the joint space.
\end{abstract}
\section{Introduction}
Planar parallel manipulators have received a lot of attention \cite{Hunt_1983, Hunt_1993,Innocenti_1992,Mcaree_1999,Kong_2001,Husty_2009,Wenger_1998,Wenger_2009,Wenger_2007,Zein_2006,Urizar_2010,Zein_2007,Bamberger_2008,Macho_2008,Hernandez_2008,Urizar_2009a,Urizar_2009b,Macho_2007}  because of their relative simplicity with respect to their spatial counterparts. Moreover, studying the former may help understand the latter. Planar manipulators with three extensible leg rods, referred to as 3-RPR manipulators, have often been studied. Such manipulators may have up to six assembly modes (AM) \cite{Hunt_1993} and their direct kinematics can be written in a polynomial of degree six \cite{Rojas_2011}. 
It was first pointed out that to move from one assembly mode to another, the manipulator should cross a singularity \cite{Hunt_1993}. However, \cite{Innocenti_1992} showed, using numerical experiments, that this statement is not true in general. More precisely, this statement is only true under some special geometric conditions, such as similar base and mobile platforms \cite{Mcaree_1999,Kong_2001}. Recently, \cite{Husty_2009} provided a mathematical proof of the decomposition of the workspace into two aspects (singularity-free regions) using geometric properties of the singularity surfaces. Since a parallel manipulator becomes uncontrollable on a singular configuration, the possibility to change its assembly-mode without encountering a singularity is interesting as it can enlarge its usable workspace. Knowing whether a parallel manipulator has this feature is of interest for both the designer and the end-user. The second-order singularities, which form cusp points in plane sections of the joint space, play an important role in non-singular assembly-mode changing motions. Indeed, encircling a cusp point makes it possible to execute such motions \cite{Mcaree_1999,Wenger_2009,Zein_2007,Bamberger_2008,Macho_2008,Hernandez_2008,Urizar_2009a,Urizar_2009b,Macho_2007,Urizar_2010} 
A special class of planar 3-RPR manipulators has been studied recently \cite{Wenger_2009,Wenger_2007}. These manipulators have the peculiarity that the resolution of the direct kinematics problem is split into a cubic equation and a quadratic equation. Their geometry is characterized by the fact that the platform triangle is congruent to the base triangle via an indirect isometry of the plane; this is the reason why we call them ``symmetric''. 

We propose here a coordinate system for the workspace which is adapted
to this specific class and reflects the splitting of the direct
kinematic problem (section II). We pay attention to the
description of singularities (section III) and cusps (section IV) using
these coordinates. We show how to sort assembly modes and use this
sorting to do motion planning in the joint space (section V).

\section{Alternative coordinates for the workspace}

The base triangle is denoted by $A_1A_2A_3$. In the direct orthonormal frame
$\mathcal{F}$
with origin $A_1$ and first axis oriented by $\overrightarrow{A_1A_2}$, the
coordinates of $A_2$ are $(b,0)$ and those of $A_3$ are $(d,h)$. The platform
triangle is denoted by $B_1B_2B_3$. Due to the symmetry property, the
coordinates of $B_2$ and $B_3$ in the direct orthonormal frame
with origin $B_1$ and first axis oriented by $\overrightarrow{B_1B_2}$ are
respectively $(b,0)$ and $(d,-h)$. The length of the leg $A_iB_i$ is as usual
denoted by $\rho_i$.  

\begin{figure}[ht!]
\centering
  \psfrag{B1}{$B_1$}
  \psfrag{B2}{$B_2$}
  \psfrag{B3}{$B_3$}
  \psfrag{A1}{$A_1$}
  \psfrag{A2}{$A_2$}
  \psfrag{A3}{$A_3$}
  \psfrag{b}{$b$}
  \psfrag{h}{$h$}
  \psfrag{d}{$d$}
  \includegraphics[scale=.75]{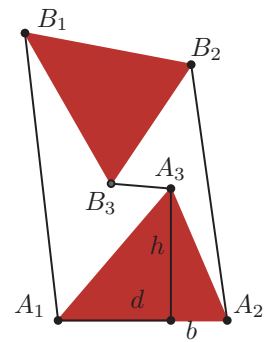}
  \caption{A symmetric 3-R\underline{P}R manipulator.\label{figparam}}
\end{figure}
\begin{figure}[h!]
\begin{center}
  \psfrag{90}{$90^{\circ}$}
  \psfrag{B1}{$B_1$}
  \psfrag{B2}{$B_2$}
  \psfrag{B3}{$B_3$}
  \psfrag{A1}{$A_1$}
  \psfrag{A2}{$A_2$}
  \psfrag{A3}{$A_3$}
  \psfrag{Aa1}{$A'_1$}
  \psfrag{Aa2}{$A'_2$}
  \psfrag{Aa3}{$A'_3$}
  \psfrag{2g}{2g}
  \psfrag{b}{$b$}
  \psfrag{h}{$h$}
  \psfrag{d}{$d$}
  \psfrag{Delta}{$\Delta$}
  \includegraphics[scale=.8]{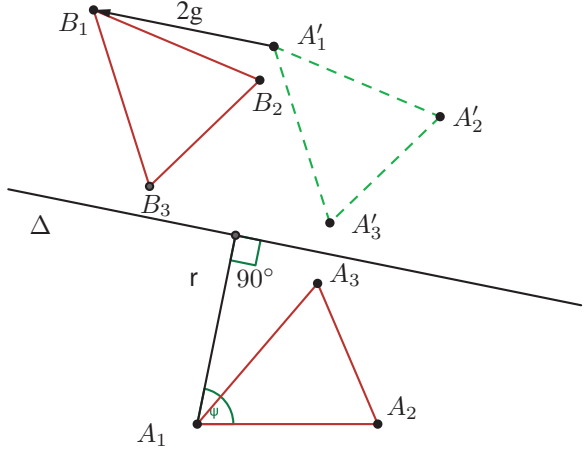}
\end{center}
\caption{The glide reflection sending $A_1A_2A_3$ to
$B_1B_2B_3$.\label{figglidedsym}}
\end{figure}

The platform triangle $B_1B_2B_3$ is the image of the base triangle $A_1A_2A_3$
by a glide reflection $S$. We encode this glide reflection $S$ by the triple
$(\psi,r,g)$ such that the glide reflection is the orthogonal symmetry with
respect to the line $\Delta$ with equation $x\,\cos(\psi) + y\,\sin(\psi) -r= 0$
followed by the
translation of vector $2g\left(\begin{array}{c} -\sin(\psi)\\ \cos(\psi)
\end{array}\right)$ parallel to the symmetry axis (the equation of $\Delta$ and
the coordinates of the translation vector are given in frame
$\mathcal{F}$ attached to the base triangle - see figure \ref{figglidedsym}). 

We
choose the angle $\psi$
in $[-\pi/2,\pi/2]$ and make
the identification of $(-\pi/2,r,g)$ with $(\pi/2,-r,-g)$.\par\medskip

Usually the workspace is viewed as the space of rigid motions in the plane and
a pose of the manipulator is encoded by the rigid motion $R$ carrying the
half-line
$[A_1A_2)$ to the half-line $[B_1B_2)$. The rigid motion $R$ and the glide
reflection $S$ are related in the following way: $S$ is the orthogonal symmetry
with
respect to $(A_1A_2)$ followed by $R$. If the rigid motion $R$ is given (as in
\cite{Wenger_2009}, for instance) by the angle of rotation $\varphi$ and the
translation vector $\left(\begin{array}{c} x\\ y
\end{array}\right)$, then the relation between the two systems of coordinates
is as follows:
\begin{eqnarray}
 \varphi &=& 2\psi+\pi\ \pmod{2\pi}\\
 x&=& 2\left(r\,\cos(\psi)-g\,\sin(\psi)\right)\\
 y&=& 2\left(r\,\sin(\psi) + g\,\cos(\psi)\right)\;.\label{chcoord}
\end{eqnarray}

It is easy to compute the lengths $\rho_i$ of the legs $A_iB_i$ in terms of
$(\psi,r,g)$, since $B_i$ is the image of $A_i$ by the glide reflection. The
square $\rho_i^2$ is the sum of the square of the double of the distance of
$A_i$ to the axis $\Delta$ and the square of the norm of the
translation vector, which is $4g^2$. This gives:
\begin{eqnarray}
 \rho_1^2 &=& 4\left(r^2+g^2\right)\;,\label{eq1}\\
\rho_2^2 &=& 4\left((b\,\cos(\psi) - r)^2 + g^2\right)\;,\label{eq2}\\
\rho_3^2 &=& 4\left((d\,\cos(\psi) + h\sin(\psi) -r)^2+
g^2\right)\;.\label{eq3}
\end{eqnarray}
It will be convenient to introduce $\delta_2= (\rho_2^2-\rho_1^2)/4$ and
$\delta_3=(\rho_3^2-\rho_1^2)/4$. These quantities depend only on $\psi$ and
$r$, and not on $g$:
\begin{eqnarray}
\!\delta_2 \!\!\!\!&=&\!\!\!\! -2\,b\cos \left( \psi \right) r+{b}^{2} \left( \cos \left( \psi
\right)  \right) ^{2}\;,\label{eq4}\\
\!\delta_3 \!\!\!\!&=&\!\!\!\! \left( d\cos \left( \psi \right)\! + \! h\sin \left( \psi
\right)  \right) ^{2}\! - \! 2 r\left(d\cos \left( \psi \right)\! + \! h\sin \left( \psi
\right)  \right)\label{eq5}
\end{eqnarray}
Eliminating $r$ between these two equations and writing the equation obtained in $t=\tan(\psi)$, we get the third degree equation:
\begin{multline}\label{eq6}
 \delta_{{2}}h{t}^{3}+ \left(
b{h}^{2}-b\delta_{{3}}+\delta_{{2}}d \right) {t}^{2}+ \\
\left(2\,bdh-{b}^{2}h+\delta_{{2}}h \right)t
-b\delta_{{3}}+\delta_{{2}}d+b{d}^{2}-{b}^{2}d=0\;. 
\end{multline}
This equation is essentially the same as the third degree characteristic
polynomial obtained in \cite{Wenger_2007}.\par\medskip

We shall use $(\rho_1^2, \rho_2^2, \rho_3^2)$ as coordinates
for the actuated joint space. Of course, this joint space is contained in the
positive
orthant $(\mathbb{R}_+)^3$.

The direct kinematic problem (DKP) can be solved as follows: 
\begin{enumerate}
 \item Take a real solution in $t$ of equation (\ref{eq6}), which
determines $\psi=\arctan(t)$. Generically there are 3 or 1 real solutions,
depending on the sign of the discriminant of the equation.
\item Compute $r$ from $\psi$ using equation (\ref{eq4}), which gives 
\begin{equation}
 r= \frac{1}{2}\,\left(b\,\cos(\psi)-
\frac{\delta_2}{b\,\cos(\psi)}\right)\;.\label{eqr}
\end{equation}
\item Solve the equation (\ref{eq1}) for $g$. It has two real opposite
solutions when $\rho_1^2 > 4r^2$.
\end{enumerate}

One should take care in the resolution of the DKP of the
case $\psi=\pm \pi/2$, i.e. $t=\infty$. This corresponds to the vanishing of
the third degree term in equation (\ref{eq6}), which occurs when $\delta_2 =
0$. In this case one can use equation (\ref{eq5}) to compute $r$, which gives
$r=(h^2-\delta_3)/2h$ for $\psi=\pi/2$. 

Note that the existence of a solution to the DKP depends
\begin{itemize}
\item first, on the existence of a solution $(\psi,r)$ to the system of
equations (7) and (8),
\item second, given such a solution $(\psi,r)$, on the existence of a solution
$g$ to equation (4).
\end{itemize}
These two conditions of existence will be important for the discussion of
singularities in the next section.
\section{Singularities}
The singular surface in the actuated joint space is thus given as the union of
two surfaces $S_1$ and $S_2$, corresponding respectively to steps 1 and 3 of
the resolution of the DKP described above. The fact that
the singular surface splits in two components has already been observed in
\cite{Wenger_2009}. We will now
describe these two surfaces. We will also describe the critical surfaces
$\Sigma_1$ and $\Sigma_2$ in the workspace, whose images by the mapping
$(\psi,r,g)\mapsto (\rho_1^2,\rho_2^2,\rho_3^2)$ given by equations
(\ref{eq1}), (\ref{eq2}) and (\ref{eq3}) are $S_1$ and $S_2$ respectively.

\subsection{The first singular surface}
The surface $S_1$ is the intersection of the actuated joint space (always with
coordinates $(\rho_1^2,\rho_2^2,\rho_3^2)$) with a cylinder
having generatrix parallel to $(1,1,1)$ and basis a curve $C$ in
the plane of coordinates $(\delta_2,\delta_3)$. An equation for $C$ can
be obtained as the discriminant of equation (\ref{eq6}); it is a quartic. An
alternative way to describe $C$ is to compute the jacobian determinant of the
mapping $\Phi:(\psi,r)\mapsto (\delta_2,\delta_3)$ given by equations
(\ref{eq4}) and (\ref{eq5}). The jacobian curve $\Gamma$ in the space with
coordinates $(\psi,r)$ is given by 
\begin{equation}
 r=\frac{\cos(\psi)}{h}\,
\left(
 \begin{array}{c}
 (h^2+b d-d^2)\,\cos(\psi)\,\sin(\psi)+\\
(2d-b)\,h\,\cos(\psi)^2 +(b-d)\,h
\end{array}
\right)
\;.
\label{rcrit}
\end{equation}
Observe that $r(\psi+\pi)=-r(\psi)$. The critical surface $\Sigma_1$ in the
workspace is the set of all $(\psi,r,g)$ such that $(\psi,r)$ belongs to
$\Gamma$. 
The curve $C$ is the image of $\Gamma$ by the mapping $\Phi$, and
it can be parameterized by rational functions of $t=\tan(\psi)$ as
\begin{equation}
 \left\{\begin{array}{rcl}
 \!\delta_2 \!\!\!&=&\!\!\!\displaystyle\frac{       
b\,((2d-b)h\,t^2+(2d^2-2bd-2h^2)\,
t+(b-2d)h)}{(1+t^2)^2\,h}\\
  \!\delta_3 \!\!\!&=&\!\!\!
\displaystyle\frac{(h\,t+d)^2\,(h\,t^2+2(d-b)\,t-h)}{(1+t^2)^2\,h}
        \end{array}\right. 
\end{equation}
So $C$ is indeed a rational quartic. Its singular points are three real cusps that can
be found by looking at the stationary points of the parameterization. These
stationary points correspond to parameters $t$ which are roots of the cubic
equation 
\begin{equation}
 (b-2d)h\,t^3+3(h^2-d^2+db)\,t^2+3h(2d-b)\,t+d^2-db-h^2=0\;.
\end{equation}
Since the discriminant
$108\,(d^2+h^2)^2\,((d-b)^2+h^2)^2$ of this cubic equation is strictly
positive, there are always three real roots and hence three real cusps on the
curve $C$. Actually, transforming the equation to an equation in $\psi$, one
obtains
\begin{equation}\label{psicusp}
 \tan(3\psi)= \frac{d^2-bd-h^2}{(b-2d)\,h}
\end{equation}
Note that the curve $C$ has no other singular
point than the three cusps. Indeed, a rational quartic may have only up to three
singular points. The curve $C$ always has the shape of a deltoid, i.e. a
closed curve with three cusps connected by arcs concave to the exterior.
\subsection{The second singular surface}
The second critical surface $\Sigma_2$ in the workspace is simply given by $g=0$; so it is independent of the geometry of the manipulator (this is already observed in \cite{Wenger_2009}).  
Its image $S_2$ in the actuated joint space is parameterized by substituting $g=0$ in equations (4-6). 
So the surface $S_2$ is the image of the elliptic cylinder
\begin{eqnarray}
 \rho_1 &=& 2\,r\;,\\
\rho_2 &=& 2\,(r -b\,\cos(\psi)) \;,\\
\rho_3 &=& 2\,(r-d\,\cos(\psi) - h\sin(\psi))\;
\end{eqnarray}
by the mapping $(\rho_1,\rho_2,\rho_3)\mapsto (\rho_1^2,\rho_2^2,\rho_3^2)$. The implicit equation of $S_2$ can also be obtained by eliminating $t$ between
 equation (\ref{eq6}) and the equation  
 $$\rho_1^2= \frac{\delta_2^2(1+t^2)}{b^2}-2\delta_2+\frac{b^2}{1+t^2}\;,$$
where the right hand side is the expression for $4r^2$ derived from (\ref{eqr}). The implicit equation for $S_2$ obtained in this way is a quartic equation in $\rho_1^2,\rho_2^2,\rho_3^2$, not a very nice one.
\subsection{An example}
We consider the manipulator with parameters $b=1$, $h=1$, $d=0$. In this case the curve $C$ in the plane $(\delta_2,\delta_3)$ is a hypocycloid with
three cusps (a deltoid) inscribed in the circle with center $(1/4,1/4)$ and radius $\sqrt{9/8}$. The three cusps on $C$ correspond to the values
$-\displaystyle\frac{5\pi}{12},\ -\displaystyle\frac{\pi}{12},\ \displaystyle\frac{\pi}{4}$ of $\psi$.

\begin{figure}[ht!]
\begin{center}
  \small{
  \psfrag{-0.4}{-0.4}
  \psfrag{-0.2}{-0.2}
  \psfrag{0.2}{0.2}
  \psfrag{0.4}{0.4}
  \psfrag{0.6}{0.6}
  \psfrag{0.8}{0.8}
  \psfrag{1}{1}
  \psfrag{1.2}{1.2}}
  \psfrag{delta2}{$\delta_2$}
  \psfrag{delta3}{$\delta_3$}
  \includegraphics[scale=.3]{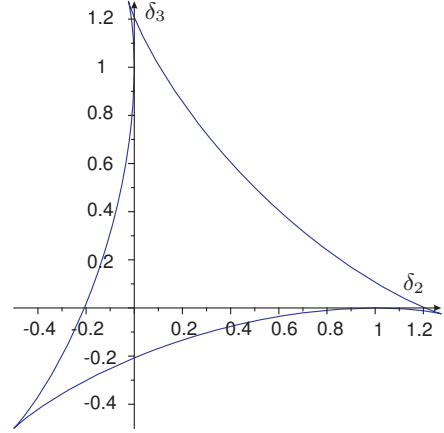}
\end{center}
\caption{The curve $C$}
\end{figure}

The critical surface $\Sigma_2$ in the workspace is always given by $g=0$. The critical surface $\Sigma_1$ is parameterized by
$$\Big(\psi,\ -\cos(\psi)\,(-1-\sin(\psi)\,\cos(\psi)+\cos(\psi)^2),\ g\Big)\;.$$

\begin{figure}[ht!]
\begin{center}
\includegraphics[scale=.4]{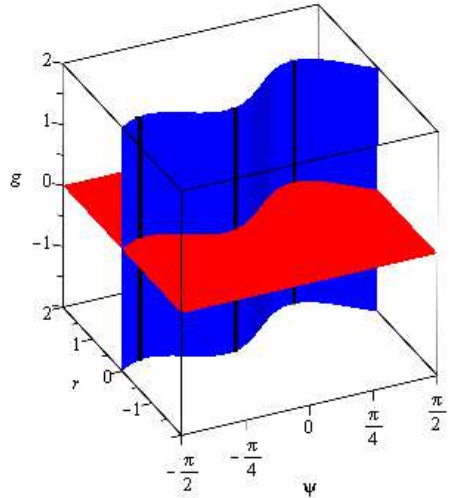}
\end{center}
\caption{The critical surfaces $\Sigma_1$ (in blue) and $\Sigma_2$ (in
red).\label{fig2}}
\end{figure}

Both critical surfaces are represented in figure \ref{fig2}. It may be interesting to compare this figure with figure 2 in \cite{Wenger_2009}, which represents the same surfaces (with the same color code), but in a different coordinate system. The choice of coordinates made here ``straightens'' the critical surfaces. 

The three black lines of figure \ref{fig2} are the lines of points which correspond to cusps in the joint space. 

\begin{figure}[ht!]
\begin{center}
\includegraphics[scale=.21]{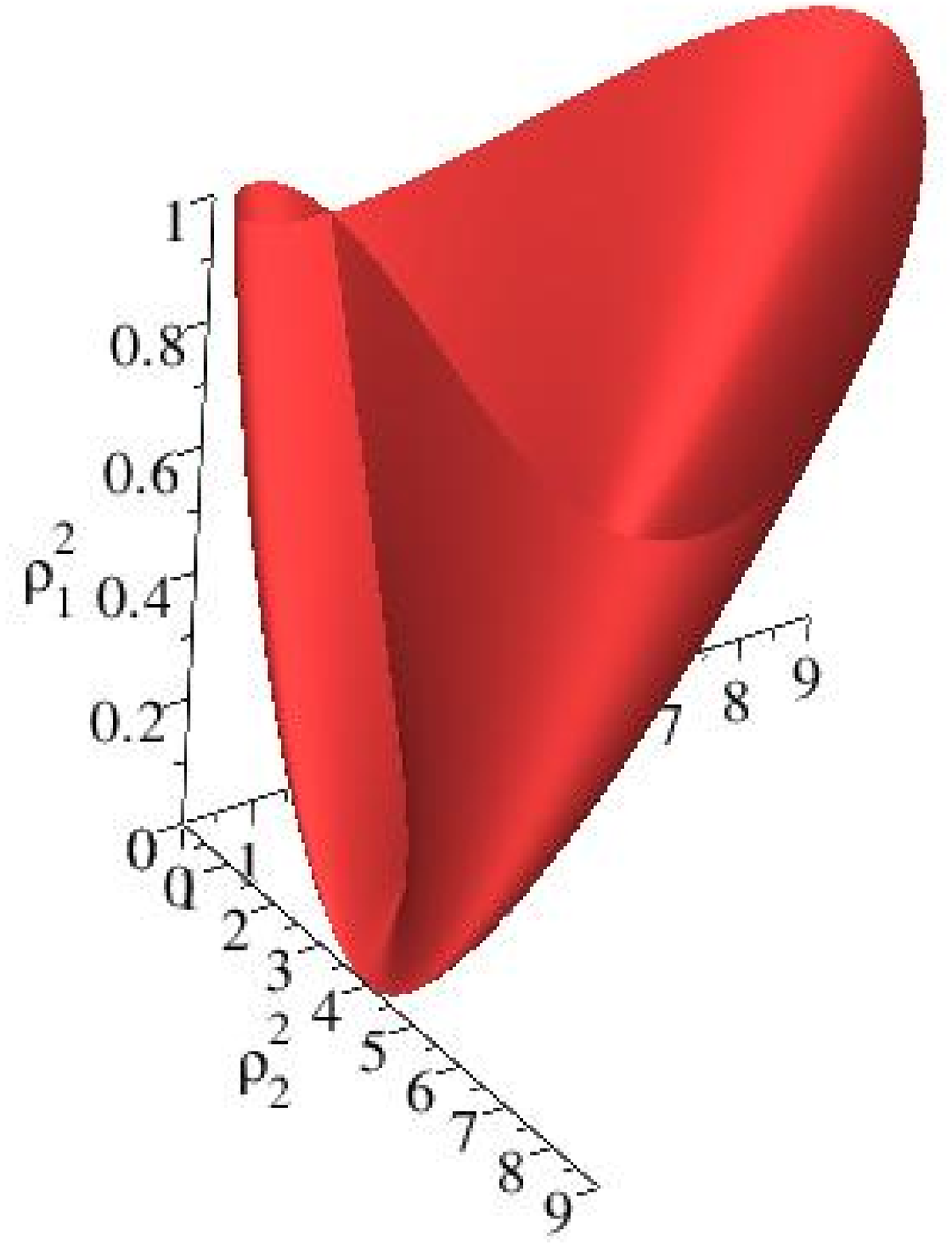}
\includegraphics[scale=.21]{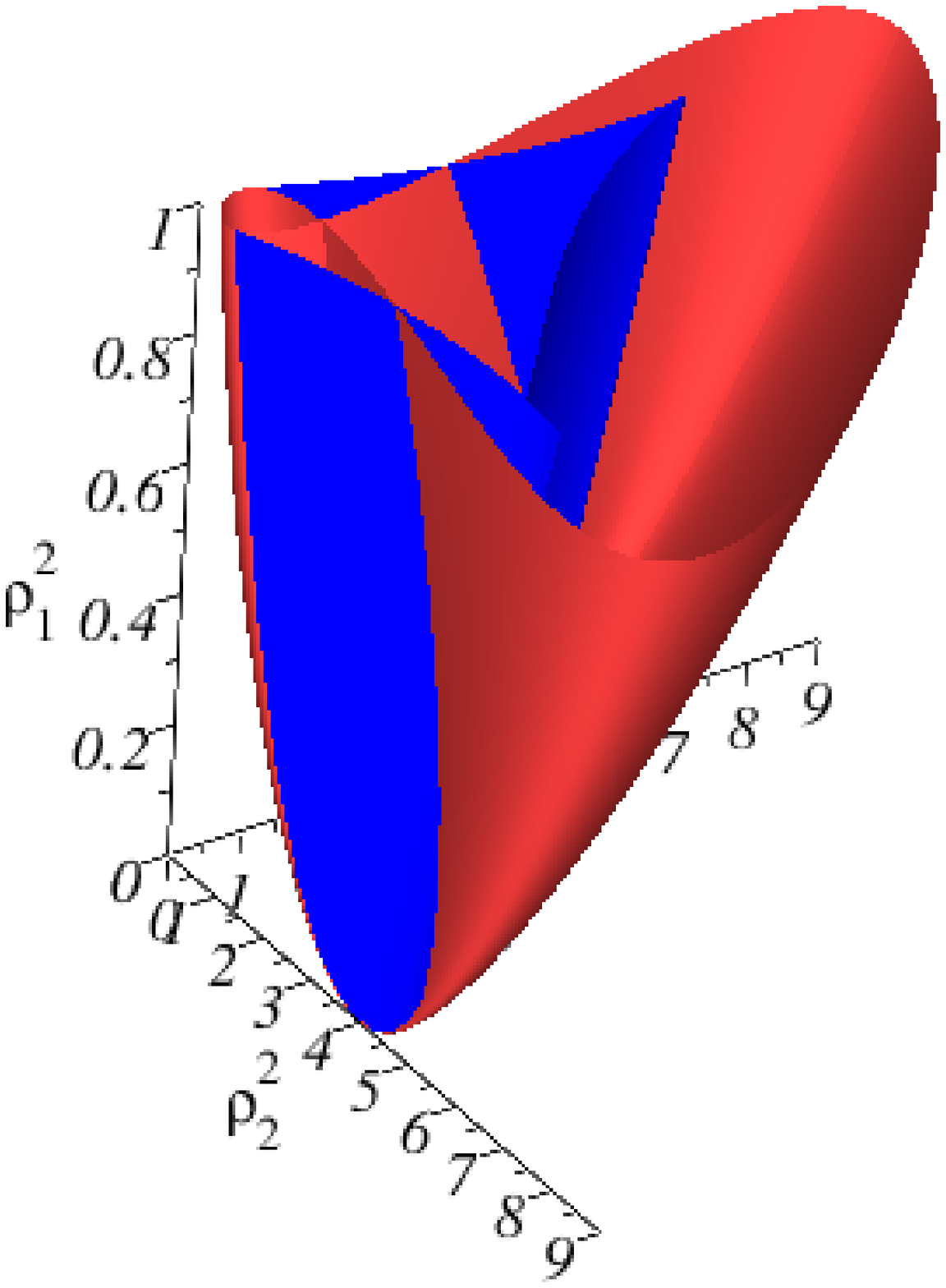}
\end{center}
\caption{$S_1$ (in blue) and $S_2$ (in
red) cut at $\rho_1^2=1$.\label{fig3}}
\end{figure}

\begin{figure}[ht!]
\begin{center}
\includegraphics[scale=.21]{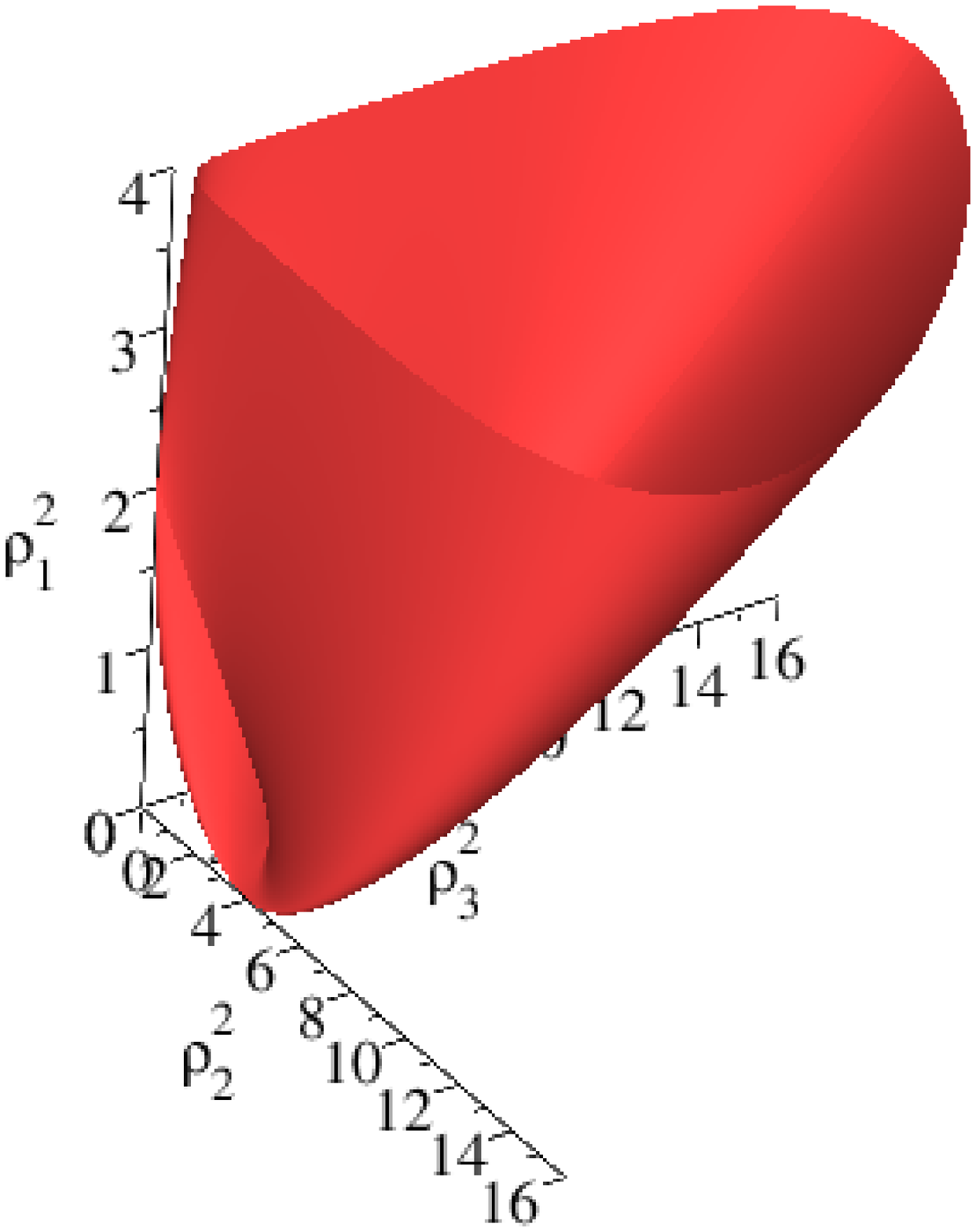}
\includegraphics[scale=.21]{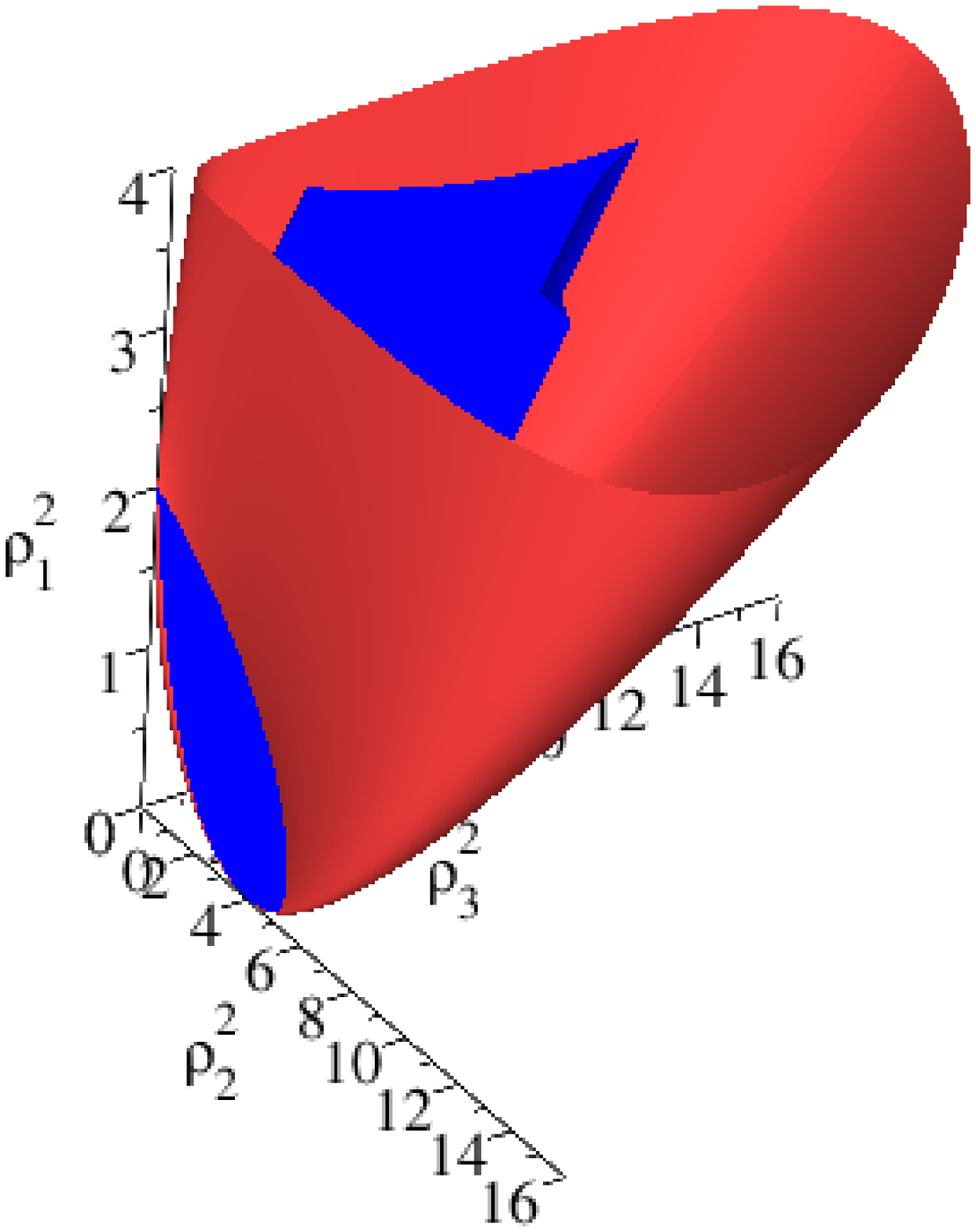}
\end{center}
\caption{$S_1$ (in blue) and $S_2$ (in
red) cut at $\rho_1^2=4$.\label{fig4}}
\end{figure}

We represent now the singular surfaces $S_1$ and $S_2$ in the joint space (See figures \ref{fig3} and \ref{fig4} ). The surface $S_1$ is a part of a cylinder on the hypocycloid and has three
half-lines of cusps. The drawing of the singular surfaces is made using their parameterizations by $(\psi,g)$ for $S_1$ and by $(\psi,r)$ for $S_2$.

\section{Cusps}

The singular surface $S_1$ has three half-lines of cusps, all parallel to the vector $(1,1,1)$. So the cusps are entirely determined once we know the origins of these three half-lines. The three possible angles $\psi$ are determined by equation (\ref{psicusp}):
$$\psi_\mathrm{cusp}= \frac{1}{3}\,
\arctan\left(\frac{d^2-bd-h^2}{(b-2d)\,h}\right)+ k\,\frac{\pi}{3}
\pmod{\pi}$$
for $k=0,1,2$.
For each of these three values of $\psi_\mathrm{cusp}$, the corresponding value $r_\mathrm{cusp}$ is given by equation (\ref{rcrit}). The couple $(\psi_\mathrm{cusp},r_\mathrm{cusp})$ determines the line supporting the half-line of cusps, and the origin of this half-line corresponds to $g=0$. So we get three values for the $\rho_1^2$ of the origins of the three half-lines of cusps, which are the three values for $4r_\mathrm{cusp}^2$. These three values for $\rho_1^2$ are the roots of a third degree polynomial with coefficients depending on $b,h,d$; the constant term of this polynomial is  $$4h^2\,(-2d+b)^2\,(d^3-2bd^2+h^2d+b^2d-2bh^2)^2\,(-h^2+2bd-d^2)^2\;,$$ and its discriminant is always nonnegative.\par\medskip

Let $0\leq \beta_1\leq \beta_2\leq \beta_3$ be the three bifurcation values of $\rho_1^2$ for the number of cusps, i.e. the $\rho_1^2$ of the origins of the half-lines of cusps. Then the slice at $\rho_1^2=c$ has 0 cusp if $0\leq c < \beta_1$, 1 if $\beta_1< c < \beta_2$, 2 if $\beta_2<c<\beta_3$ and 3 if $\beta_3<c$. One of the bounded intervals may be empty, if the constant term of the equation of the third degree in $\rho_1^2$ vanishes (for the first interval) or if its discriminant  vanishes (for the second or third interval). One has to understand that the number of cusps is the number of cusp points in the slice of the joint space. Over each of these cusps there are two triple solutions of the DKP, corresponding to opposite values of
$g$.\par\medskip

\begin{figure}[ht!]
\begin{center}
\includegraphics[scale=.4]{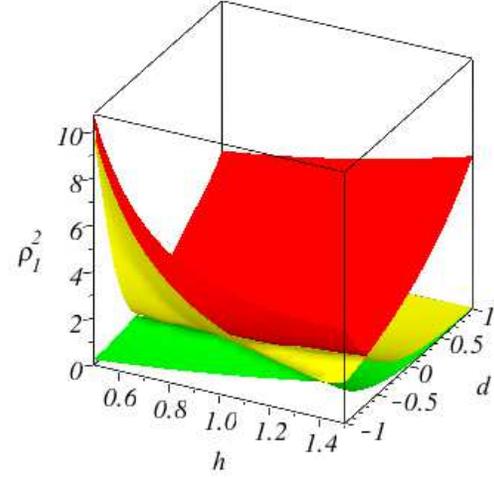}\goodbreak(a)\goodbreak
\includegraphics[scale=.4]{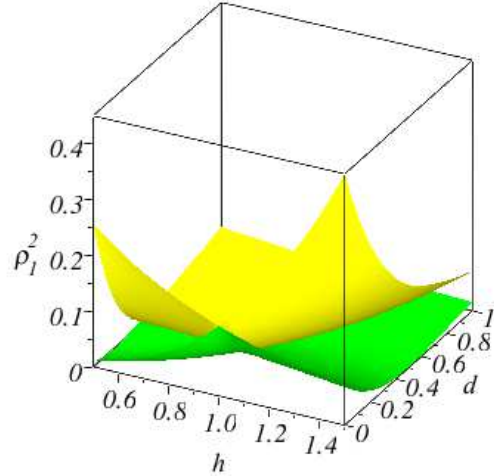}\goodbreak(b)
\end{center}
\caption{The bifurcation values of $\rho_1^2$
for the number of cusps.\label{figbif}}
\end{figure}

Figure \ref{figbif}a represents the bifurcation values as function of $h$ and $d$, with $b$ fixed equal to 1 (green for $\beta_1$, yellow for $\beta_2$ and red for $\beta_3$). Figure \ref{figbif}b shows detail for the first and second bifurcation values when $0\leq d\leq 1$.
\section{Sorting assembly modes and motion planning in the joint space}
The essential idea here is the following: when one starts from a nonsingular solution of the DKP at a point in the joint space with coordinates $(\rho_1^2, \rho_2^2, \rho_3^2)$ and moves in the direction of the vector $(1,1,1)$, then the solution of the DKP follows smoothly, without crossing a singularity in the workspace. Indeed, consider equations (4-6): the motion $\rho_i^2\mapsto \rho_i^2+\lambda ^2$, increasing $\lambda$, can be lifted in the workspace (with coordinates $(\psi,r,g)$) to a path with $\psi$ and $r$ fixed and $g$ increasing to $g^2+\lambda^2/4$. 

The segment in the joint space can cross the second singular surface $S_2$. This corresponds to the appearance of two new solutions to the DKP (one in each
aspect), with a different couple $(\psi,r)$. But it never crosses the first singular surface $S_1$ which is a cylinder with generatrix parallel to
$(1,1,1)$. 

We denote $\rho_1^2+\rho_2^2+\rho_3^2$ by $\nu$. Then $(\nu,\delta_2,\delta_3)$ form a system of coordinates for the joint space which is convenient for our
present discussion. Moving in the joint space in the direction of the vector $(1,1,1)$ is increasing $\nu$, keeping $\delta_2$ and $\delta_3$ fixed.
\par 
\begin{figure}[ht!]
\begin{center}
\psfrag{psi2}{$\delta_2$}
\psfrag{psi3}{$\delta_3$}
\psfrag{-1}[c]{$-1$}
\psfrag{-0.5}[c]{$-0.5$}
\psfrag{0}[c]{$0$}
\psfrag{0.5}[c]{$0.5$}
\psfrag{1}[c]{$1$}
\psfrag{1.5}[c]{$1.5$}
\psfrag{2}{$2$}
\psfrag{3}{$3$}
\includegraphics[scale=.27]{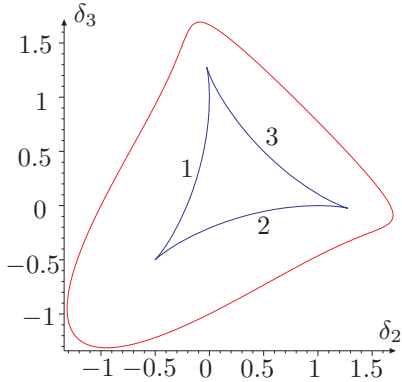}
\end{center}
\caption{The section $\nu=8$.\label{figSSS8}}
\end{figure}
The situation in the section $\nu=k$ stabilizes for sufficiently large values of $k$: the section of $S_2$ is a big oval surrounding the section of $S_1$ which is a deltoid with three cusps (the curve $C$, base of the cylinder). Inside this curve $C$ there are, in each aspect, three continuous solutions of the DKP and between this curve $C$ and the section of $S_2$ there is in the same aspect one continuous solution of the DKP. Label by 0 this solution, and label by $1, 2, 3$ the three arcs of $C$ between the cusps. Then we can label the three solutions inside the deltoid $C$ by $1, 2, 3$ according to the label of the arc of the deltoid through which they are connected with the solution 0. Figure \ref{figSSS8} illustrates the labelling in the same example as above (the coordinates $(\delta_2,\delta_3)$ are used in the section $\nu=8$). 

In this way we can label every solution of the DKP contained in one aspect by one of the labels 0,1,2 or 3. In each aspect, all points in the same label form a connected region and the boundaries between these regions are the so-called ``characteristic surfaces'' obtained by pulling back the singular surface $S_1$ in the aspect \cite{Wenger_1997}. The characteristic surfaces in the workspace with coordinates $(\psi,r,g)$ are cylinders with generatrix parallel to the $g$-axis and basis the two curves in the $(\psi,r)$ plane which are given by  
\begin{multline*}
r = \frac{h(2\,d+b)\,\cos \left( \psi \right) + ({h}^{2}+bd-{d}^{2})\,\sin \left( \psi
\right)}{4\,h} \\
\pm \frac{\sqrt{ \left( {h}^{2}+{d}^{2} \right)  \left(
{h}^{2}+{d}^{2}+{b}^{2}-2\,bd \right)}}{4\,h}
\end{multline*}

\begin{figure}[ht!]
\begin{center}
  \small{
  \psfrag{-0.6}{$-0.6$}
  \psfrag{-0.4}{$-0.4$}
  \psfrag{-0.2}{$-0.2$}
  \psfrag{0}{$0$}
  \psfrag{0.2}{$0.2$}
  \psfrag{0.4}{$0.4$}
  \psfrag{0.6}{$0.6$}
  }
  \psfrag{r}{$r$}
  \psfrag{phi}{$\psi$}
  \psfrag{0r}{\color[rgb]{1,0,0}0}
  \psfrag{1r}{\color[rgb]{1,0,0}1}
  \psfrag{2r}{\color[rgb]{1,0,0}2}
  \psfrag{3r}{\color[rgb]{1,0,0}3}
  \psfrag{0b}{\color[rgb]{0,1,0}0}
  \psfrag{1b}{\color[rgb]{0,1,0}1}
  \psfrag{2b}{\color[rgb]{0,1,0}2}
  \psfrag{3b}{\color[rgb]{0,1,0}3}
  \psfrag{p1}{$\frac{\pi}{4}$}
  \psfrag{p2}{$\frac{\pi}{2}$}
  \psfrag{p3}{$\frac{3\pi}{4}$}
  \psfrag{p4}{$\pi$}
  \psfrag{p5}{$\frac{5\pi}{4}$}
  \psfrag{p6}{$\frac{3\pi}{2}$}
  \psfrag{p7}{$\frac{7\pi}{4}$}
  \psfrag{p8}{$2\pi$}
  \includegraphics[scale=.45]{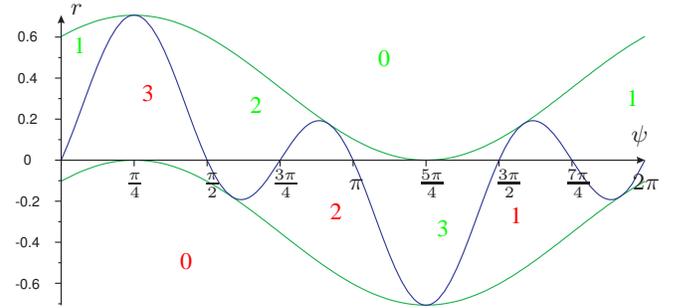}
\end{center}
\caption{The four sorts of assembly modes in each aspect, in a section $g=\mbox{constant}$ of the workspace.  The green curves are sections of the characteristic surfaces. \label{figchar}}
\end{figure}

Figure \ref{figchar} represents a section of the characteristic surfaces (in green) by a plane $g=\mbox{constant}$ of the workspace. The blue curve is a
section of $\Sigma_1$ and separates the aspects. In each aspect, the characteristic surface delimitates the four sorts of assembly modes. The label 0 correspond to points which are mapped outside of the deltoid $C$ in the joint space, and the labels $1, 2, 3$ to points which are mapped inside. A path from 0 to 1 inside an aspect is mapped to a path going through the arc of the deltoid with label 1, etc..

\par
\begin{figure}[ht!]
\begin{center}
  \psfrag{B1}{$B_1$}
  \psfrag{B2}{$B_2$}
  \psfrag{B3}{$B_3$}
  \psfrag{A1}{$A_1$}
  \psfrag{A2}{$A_2$}
  \psfrag{A3}{$A_3$}
  \psfrag{nu}{$\nu\approx7.99$}
  \psfrag{delta2}{$\delta_2\approx-1.06$}
  \psfrag{delta3}{$\delta_3\approx-1.06$}
  \psfrag{2g}{2g}
  \psfrag{r}{r}
  \psfrag{rr}{$r=1.1$}
  \psfrag{g}{$g=0.4$}
  \psfrag{phi}{$\psi=45^{\circ}$}
  \psfrag{1.12}{1.12}
  \psfrag{2.34}{2.34}
\includegraphics[scale=.7]{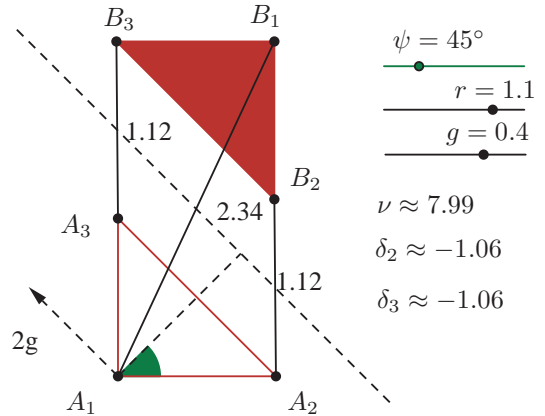}
\end{center}
\caption{The goal position. \label{figIKhome}}
\end{figure}

\medskip
We illustrate how the labelling can be used for motion planning in the joint space with an example, again for the manipulator with parameters $b=1$, $h=1$, $d=0$. We choose a goal position for the manipulator, given by $\psi= \pi/4$, $r=1.1$ and $g=0.4$ (Figure \ref{figIKhome}).

\begin{figure}[ht!]
\begin{center}
\psfrag{psi2}{$\delta_2$}
\psfrag{psi3}{$\delta_3$}
\psfrag{-1}[c]{$-1$}
\psfrag{-0.5}[c]{$-0.5$}
\psfrag{0}[c]{$0$}
\psfrag{0.5}[c]{$0.5$}
\psfrag{1}[c]{$1$}
\psfrag{1.5}[c]{$1.5$}
\psfrag{2}{$2$}
\psfrag{3}{$3$}
\psfrag{home}{Goal}
\psfrag{start}{Start}
\includegraphics[scale=.27]{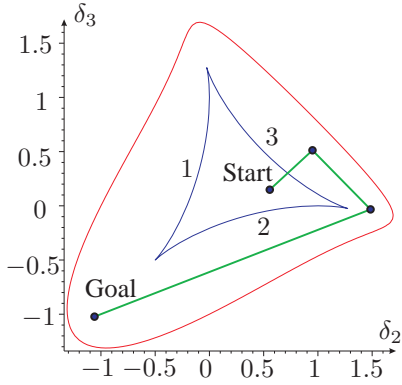}
\caption{A path from a starting and a goal position\label{path}}
\end{center}
\end{figure}

The goal position corresponds to values $\rho_1 \approx 2.34$, $\rho_2=\rho_3\approx 1.12$ and $\nu\approx 7.99$; it is labelled 0 and mapped outside of the deltoid in the joint space.
We explain how to plan (in the joint space) a path to the goal position from a position of the manipulator in the same aspect, with $\nu\leq 7.99$. The starting position corresponds to
$\rho_1^0,\rho_2^0,\rho_3^0$ and label $\ell \in\{0,1,2,3\}$.
\begin{itemize}
 \item Increase simultaneously $\rho_1,\rho_2,\rho_3$ following $\rho_i=\sqrt{(\rho_i^0)^2+t^2}$ until $\nu=\sum\rho_i^2= 8$
 \item Keeping $\nu=8$ constant, move in the plane with coordinates $(\delta_2,\delta_3)$ from 
$$(\delta_2^0=(\rho_2^0)^2-(\rho_1^0)^2,\
\delta_3^0=(\rho_3^0)^2-(\rho_1^0)^2)$$
to $(-1.06,-1.06)$ following a path inside the red curve and crossing only  arc $\#\ell$ of the deltoid if the label is $\ell$. 
\end{itemize}
Figure \ref{path} 
shows such a path for label $\ell=3$. (Actually, it shows only the part of the path in the
plane $(\delta_2,\delta_3)$ for $\nu=8$, since the first segment of the path
increases $\nu$ without changing $\delta_2$ nor $\delta_3$).

\section{Conclusions}
The choice of coordinates for the workspace well adapted to the special class of symmetric manipulators allowed us to take full advantage of the de-coupling of DKP (into a cubic and a quadratic equation) in the computations. We obtained rational parameterizations of the singular surfaces in the joint space. We obtained also a good description of the cusp curves on these surfaces, as well as precise information on the bifurcation of the number of cusps in the slices of the joint space. Finally, we were able to sort assembly modes in an aspect and to use this sorting for motion planning in the joint space. 

\end{document}